\pdfoutput=1

\documentclass[11pt]{article}

\usepackage{acl}

\usepackage{times}
\usepackage{latexsym}

\usepackage[T1]{fontenc}

\usepackage[utf8]{inputenc}

\usepackage{microtype}
\usepackage{algorithm}
\usepackage{enumerate}
\usepackage{enumitem}
\usepackage{multirow}
\usepackage{graphicx}
\usepackage{array}
\usepackage{booktabs}
\usepackage{amsmath}
\usepackage{tikz}
\usetikzlibrary{patterns}
\usepackage{pgf-pie}
\usepackage{colortbl}
\usepackage{array}   
\usepackage{fdsymbol}
\usepackage{inconsolata}
\usepackage{verbatim}
\usepackage{stmaryrd}
\usepackage{trimclip}

\newcommand{\ides}{IDS}

\usepackage{pifont}
\newcommand{\okmark}{{\textbf{\textcolor[rgb]{0.1, 0.5, 0.1}{$\checkmark$}}}}
\newcommand{\ngmark}{{\textbf{\color{red}{\ding{55}}}}}
\definecolor{applegreen}{rgb}{0.56, 0.8, 0.25}
\definecolor{mypurple}{rgb}{0.62,0.24,0.81}
\usepackage{algpseudocode}
\newcommand{\ie}{{\em i.e.,}}
\newcommand{\eg}{{\em e.g.,}}
\newcommand{\Ni}{({\em i})~}
\newcommand{\Nii}{({\em ii})~}
\newcommand{\Niii}{({\em iii})~}

\newcommand{\blue}[1]{\textcolor{blue}{#1}}

\usepackage[nameinlink]{cleveref}
\crefformat{section}{\S#2#1#3} 
\crefname{algorithm}{Alg.}{Algs.}
\crefformat{subsection}{\S#2#1#3}
\Crefname{equation}{Eq.}{Eqs.}
\Crefname{figure}{Fig.}{Figs.}

\algdef{SE}[DOWHILE]{Do}{doWhile}{\algorithmicdo}[1]{\algorithmicwhile\ #1}%
\title{In-Context Learning with Iterative Demonstration Selection}

\author{Chengwei Qin$^\dagger$\thanks{Correspondence to Chengwei Qin <chengwei003@e.ntu.edu.sg> and Aston Zhang <az@astonzhang.com>}, Aston Zhang$^\clubsuit$, Chen Chen$^\dagger$, Anirudh Dagar$^\clubsuit$,Wenming Ye$^\clubsuit$
\\
$^\dagger$Nanyang Technological University,$^\clubsuit$Amazon Web Services}

\begin{document}
\maketitle
\begin{abstract}
Spurred by advancements in scale, large language models (LLMs) have demonstrated strong few-shot learning ability via in-context learning (ICL). However, the performance of ICL has been shown to be highly sensitive to the selection of few-shot demonstrations. Selecting the most suitable examples as context remains an ongoing challenge and an open problem. Existing literature has highlighted the importance of selecting examples that are diverse or semantically similar to the test sample while ignoring the fact that the optimal selection dimension, \ie\ diversity or similarity, is task-specific. Based on how the test sample is answered, we propose Iterative Demonstration Selection (\ides) to leverage the merits of both dimensions.
Using zero-shot chain-of-thought reasoning (Zero-shot-CoT), \ides\ iteratively selects examples that are diverse but still strongly correlated with the test sample as ICL demonstrations. Specifically, \ides\ applies Zero-shot-CoT to the test sample before demonstration selection. The output reasoning path is then used to choose demonstrations that are prepended to the test sample for inference. The generated answer is followed by its corresponding reasoning path for extracting a new set of demonstrations in the next iteration. After several iterations, \ides\ adopts majority voting to obtain the final result. Through extensive experiments on tasks including reasoning, question answering, and topic classification, we demonstrate that \ides\ can consistently outperform existing ICL demonstration selection methods.
\end{abstract}

\section{Introduction}

\begin{figure}[t]
    \centering
    \includegraphics[width=0.45\textwidth]{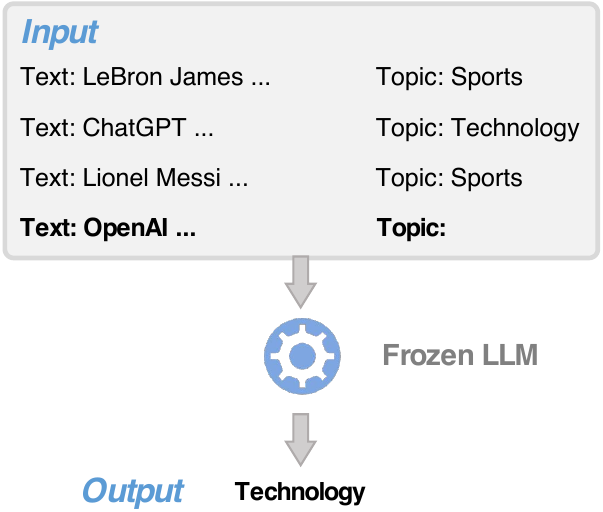}
    \caption{ Illustration of in-context learning (ICL) on topic classification. A frozen large language model directly generates the topic `Technology' for the test sample `OpenAI ...' by taking the demonstrations and the test sample as input.}
    \label{icl}
\end{figure}

With the recent advancements in scaling up model parameters, large language models (LLMs) showcase promising results on a variety of few-shot tasks through in-context learning (ICL), where the model is expected to directly generate the output of the test sample without updating parameters. This is achieved by conditioning on a manually designed prompt consisting of an optional task description and a few demonstration examples \citep{brown2020language}. \Cref{icl} shows an example describing how LLMs perform ICL on the topic classification task. Given a few text-topic pairs as demonstrations, ICL combines them with the test sample as input, to the LLM for inference. The output, \ie\ `Technology', is generated by the model autoregressively without any parameter updates.

Despite the effectiveness, the performance of ICL has been shown to be highly sensitive to the selection of demonstration examples \citep{zhao21c}. Different sets of demonstrations can yield performance ranging from nearly random to comparable with state-of-the-art models \citep{gao-etal-2021-making,lu-etal-2022-fantastically,qin2023improving}. To alleviate the above issue, researchers in ICL have proposed a number of methods to select a set of examples as few-shot demonstrations \citep{rubin-etal-2022-learning,liu-etal-2022-makes,li2023finding,wang2023large,li2023unified,ma2023fairness,an2023skill}. However, for LLMs for which parameters or detailed output distributions are not available \citep{sun2022black}, it is still a common practice to randomly select examples or select examples that are semantically similar to the test sample as demonstrations, \ie\ considering diversity or similarity.
While several approaches investigate the combination of similarity and diversity when prompting with explanations, exploring compositional generalization, or choosing examples for annotation \citep{ye-etal-2023-complementary,an-etal-2023-context,su2023selective}, 
it is not yet clear how to determine and leverage the optimal dimension for different tasks in ICL and how the rationale for answering the query benefits the balance between these two dimensions.

Actually, the optimal dimension for selecting demonstration examples is task-specific. As we will show in \Cref{sec:two_dimensions}, the diversity dimension is superior to the similarity dimension on CommonsenseQA while the similarity dimension outperforms the diversity dimension on AGNews and BoolQ. Thus, it is unreasonable to claim that one dimension is consistently better than the other across different tasks. To fully leverage the merits of both dimensions, we propose Iterative Demonstration Selection (\ides) for ICL (\Cref{fig:method}) by utilizing \emph{how the test sample is answered}. \ides\ can iteratively select demonstration examples that are diverse but still have a strong correlation with the test sample through zero-shot chain-of-thought reasoning (Zero-shot-CoT) \citep{kojima2022large}. Specifically, Zero-shot-CoT, \eg\ ``Let's think step by step.'', is first applied to the test sample before selecting demonstrations to obtain a reasoning path. The training examples that are most semantically similar to the generated reasoning path are then selected as demonstrations. They are prepended to the test sample for inference. Note that \ides\ ensures that the generated answer is accompanied by the reasoning path through designed prompts. The new reasoning path is then used for extracting another set of demonstration examples by semantic similarity in the next iteration. After a few iterations, \ides\ adopts majority voting to obtain the final result. 
Empirical results on tasks spanning mathematical reasoning, commonsense reasoning, logical reasoning, question answering, and topic classification show that \ides\ can consistently outperform previous ICL demonstration selection baselines. In summary, our main contributions are:  
\begin{itemize}[leftmargin=*]
    \item We consider both the diversity and similarity dimensions of ICL demonstration selection for LLMs. We identify that the optimal dimension for selecting demonstrations is task-specific and propose Iterative Demonstration Selection (\ides) based on how the test query is answered to fully leverage the merits of both dimensions.
    \item With extensive experiments and analysis, we demonstrate the effectiveness of \ides\ on a variety of tasks.
\end{itemize}

\section{Related Work}

This work mainly explores how to select few-shot in-context learning demonstrations for LLMs by leveraging Zero-shot-CoT. In light of this, we review four lines of research that form the basis of this work: few-shot learning, in-context learning basics, demonstration selection for in-context learning, and chain-of-thought reasoning. 

\subsection{Few-shot Learning}

Few-shot learning aims to learn tasks with only a few labeled samples, which results in a big challenge, \ie\ over-fitting, for models as they typically require large amounts of data for training. Prior methods to address over-fitting mainly focused on augmenting the few-shot data \citep{gao2020neural,qin-joty-2022-continual,qin2022lfpt}, reducing the hypothesis space \citep{triantafillou2017few,hu-etal-2018-shot,qin-etal-2023-learning}, or optimizing the strategy for searching the best hypothesis \citep{ravi2016optimization,finn2017model,xia2024chain}. More recently, LLMs have demonstrated strong few-shot learning ability through in-context learning without any parameter updates \citep{brown2020language}.

\subsection{In-context Learning} 

\citet{brown2020language} first showed that a frozen GPT-3 model can achieve impressive results on a variety of few-shot NLP tasks through conditioning on manually designed prompts consisting of task descriptions and several demonstration examples. Since then many efforts have been made on in-context learning (ICL) \citep{dong2022survey}. \citet{chen-etal-2022-improving,min-etal-2022-metaicl,wei2023symbol} demonstrated that the ICL ability of language models can be further improved through self-supervised or supervised training. Some analytical studies attempted to understand what factors affect ICL performance \citep{zhao21c,shin-etal-2022-effect,wei2022emergent,min-etal-2022-rethinking,yoo-etal-2022-ground,wei2023larger} and why ICL works \citep{xie2022an,olsson2022context,li2023transformers,pan2023context,dai2023can}. Other ongoing research on ICL has also explored \Ni demonstration designing, including demonstration selection \citep{liu-etal-2022-makes,rubin-etal-2022-learning,wang2023large}, demonstration ordering \citep{lu-etal-2022-fantastically}, and demonstration formatting \citep{cot_wei,wang2022self,zhou2023large,zhang2022automatic}, \Nii applications of ICL \citep{ding2022gpt,meade2023using,zheng2023can}, and \Niii ICL beyond text \citep{wang2023context,huang2023language,zhu2023minigpt,wang2023neural}. 

\subsection{Demonstration Selection for In-context Learning}

The performance of ICL has been shown to be highly sensitive to the selection of demonstration examples \citep{zhao21c}. Existing methods to solve this problem can be mainly divided into two categories. First, \textit{unsupervised} methods rely on pre-defined metrics. \citet{liu-etal-2022-makes} proposed to select the closest neighbors as demonstrations. In contrast, \citet{levy2022diverse} selected diverse demonstrations to improve in-context compositional generalization. More recent studies have explored leveraging the output distributions or predictive uncertainty of language models to select few-shot demonstrations \citep{wu2022self,nguyen2023context,li2023finding,ma2023fairness,ling2024uncertainty,xu2024misconfidence} or self-generating demonstrations \citep{chen-etal-2023-self}. Second, \textit{supervised} methods involve model training. \citet{rubin-etal-2022-learning,ye2023compositional,li2023unified,luo2023dr,wang-etal-2024-learning} proposed to learn to retrieve demonstration examples. \citet{wang2023large} posited LMs as implicit topic models to facilitate demonstration selection. In addition, some studies \citep{zhang2022active,scarlatos2023reticl} attempted to select demonstrations based on reinforcement learning. 
However, it is still a common practice to randomly select examples or select examples that are semantically similar to the test sample as demonstrations for LLMs for which parameters or detailed output distributions are not available \citep{sun2022black}. Several methods investigated the combination of diversity and similarity in different scenarios, \eg\ prompting with explanations \citep{ye-etal-2023-complementary}, choosing examples for annotation \citep{su2023selective} and exploring compositional generalization \citep{an-etal-2023-context}. Nevertheless, it remains unclear to us how to determine and leverage the optimal dimension for different tasks in ICL and how the reason for answering the test sample benefits the balance between the two dimensions, which motivates us to propose our simple but effective approach (\ides).

\subsection{Chain-of-Thought Reasoning}

Chain-of-thought (CoT) reasoning induces LLMs to produce intermediate reasoning steps before generating the final answer \citep{cot_wei}. Depending on whether there are manually designed demonstrations, current CoT reasoning methods mainly include Manual-CoT and Zero-shot-CoT. In Manual-CoT, human-labeled reasoning paths are used to perform CoT reasoning \citep{cot_wei,zhou2022least,cot_wei_sc,li2022advance,wang2022rationale,zhao-etal-2023-verify}. In contrast, LLMs leverage self-generated rationales for reasoning in Zero-shot-CoT \citep{kojima2022large,zelikman2022star,zhang2022automatic,diao2023active,qin2024relevant}. The ongoing  research on CoT reasoning has also explored \Ni multimodal reasoning \citep{zhang2023multicot,wu2023visual}, \Nii distilling knowledge from LLMs \citep{ho2022large,fu2023specializing}, and \Niii iterative optimization \citep{shinn2023reflexion,madaan2023selfrefine,paul2023refiner,jiao2024learning}.
\section{Problem Formulation}

Given the test set $\mathcal{D}_{\text{test}}$ and the training set $\mathcal{D}_{\text{train}}$, the goal of ICL demonstration selection is to find an optimal subset $\mathcal{S} = \{(x_1,y_1),...,(x_k,y_k)\}$ ($k$-shot) of $\mathcal{D}_{\text{train}}$ as demonstration examples for each test sample $(\hat x_i,\hat y_i)$ to maximize the overall task performance on $\mathcal{D}_{\text{test}}$. More formally, the optimal selection method $\tilde h$
is defined as:

\begin{align}
\tilde h = \mathop{\arg\max}\limits_{h \in \mathcal{H}} \sum_{i=1}^{|\mathcal{D}_{\text{test}}|} \delta_{\text{LLM}([h(\mathcal{D}_{\text{train}}, \hat x_i,\hat y_i), \hat x_i]),\hat y_i}  \label{eq:problem}
\end{align}
where $\mathcal{H}$ is the hypothesis space for searching demonstration examples, $h(\mathcal{D}_{\text{train}}, \hat x_i,\hat y_i)$ refers to demonstrations selected for $(\hat x_i,\hat y_i)$ using $h$, $[,]$ stands for concatenation, and $\delta_{a,b}$ is the Kronecker delta function: $\delta_{a,b} = 1$ if $a$ equals $b$, otherwise $\delta_{a,b} = 0$. In this work, we aim to find the optimal method $\tilde h$ by leveraging Zero-shot-CoT.
\section{What Makes Good In-Context Demonstrations?} \label{sec:two_dimensions}

\begin{table}[t]

\centering
\small
\setlength\tabcolsep{3pt}
\scalebox{0.76}{
\begin{tabular}{l|c|c|c}
\toprule
\multirow{1}{*}{} & \multicolumn{1}{c}{CommonsenseQA}
& \multicolumn{1}{|c}{BoolQ}  & \multicolumn{1}{|c}{AGNews} \\
\midrule

Similar-ICL-Consistency (\textbf{Similarity})  & 76.0 & \textbf{85.0} & \textbf{90.0} \\
Random-ICL-Voting (\textbf{Diversity}) & \textbf{79.0} & 84.0 & 88.0 \\

\bottomrule
\end{tabular}
}
\caption{Results of different methods on CommonsenseQA, BoolQ and AGNews. The optimal dimension for selecting ICL demonstrations is task-specific.} 
\label{table:pilot_exp}
\end{table}

\begin{table*}[t]

\centering
\footnotesize
\begin{tabular}{p{0.47\textwidth}|p{0.47\textwidth}}
\toprule

\textbf{Similar-ICL-Consistency} &
\textbf{Random-ICL-Voting}
\\

\midrule

Which choice is the correct answer to the question?\newline
&
Which choice is the correct answer to the question?\newline
\\
\textbf{Examples}:

\textbf{Question}: If you have cleaned off dust here it may be difficult to do your homework where? Answer Choices: (A) desktop (B) closet (C) most buildings (D) surface of earth (E) stove

\textbf{Answer}: A

\textbf{Question}: Where is dust likely to be under? Answer Choices: (A) closet (B) ground (C) windowsill (D) attic (E) carpet

\textbf{Answer}: E

\textbf{Question}: Where would you find a dustbin that is being used? Answer Choices: (A) utility closet (B) ground (C) cupboard (D) broom closet (E) kitchen

\textbf{Answer}: E

\textbf{Question}: Dust accumulates where? Answer Choices: (A) ceiling (B) library (C) surface of earth (D) \underline{\textbf{\textcolor{blue}{most buildings}}} (E) desktop

\textbf{Answer}: D
&
\textbf{Examples}:

\textbf{Question}: She had a busy schedule, she had to run errands and pick up the kids the second she did what? Answer Choices: (A) make time for (B) take money (C) go outdoors (D) leave work (E) field

\textbf{Answer}: D

\textbf{Question}: What is the worst outcome of an injury? Answer Choices: (A) cause death (B) cause bleeding (C) falling down (D) become infected (E) claim insurance

\textbf{Answer}: A

\textbf{Question}: Mom said that Sarah should stay in bed until she was able to go to school again. What did mom say to Sarah when she tried to get up? Answer Choices: (A) you're sick (B) were sick (C) more rest (D) rest more (E) get back under the covers

\textbf{Answer}: A

\textbf{Question}: John got a raise, but he lost rank. Overall, it was a good what? Answer Choices: (A) demotion (B) push down (C) go off strike (D) lower (E) go off strike

\textbf{Answer}: A
\\
The response should follow the format: Answer: \{A, B, C, D or E\}
&
The response should follow the format: Answer: \{A, B, C, D or E\}
\\
Here is the test data.
&
Here is the test data.
\\
\textbf{Question}: John wanted to clean all of the dust out of his place before settling down to watch his favorite shows.  What might he hardest do dust? Answer Choices: (A) closet (B) under the bed (C) television (D) attic (E) \underline{\textbf{\textcolor{blue}{most buildings}}}
&
\textbf{Question}: John wanted to clean all of the dust out of his place before settling down to watch his favorite shows.  What might he hardest do dust? Answer Choices: (A) closet (B) under the bed (C) television (D) attic (E) most buildings
\\
\midrule
\textbf{Answer}: \textcolor{red}{E} \ngmark
&
\textbf{Answer}: \textcolor{applegreen}{D} \okmark
\\
\bottomrule
\end{tabular}
\caption{Examples of Similar-ICL-Consistency (first decoding path) and Random-ICL-Voting (first run) for constructing demonstration examples. The upper part is the input to LLMs, including few-shot demonstrations, and the lower part is the predicted answer. Similar-ICL-Consistency gives the wrong answer `most buildings' which is actually the output of the final demonstration example, indicating that the decision process of the model is misled by this similar sample.}
\label{tab:sim_error}
\end{table*}

As demonstrated in previous work \citep{zhao21c}, the overall task performance is highly sensitive to the selection method $h$. Different sets of demonstration examples can yield significantly different performance. For example, \citet{zhang2022active} show that the minimum and maximum ICL performance due to random sampling differs by $> 30\%$ on 4 classification tasks, which emphasizes the importance of selecting good demonstrations for LLMs. 

A natural question is: what makes good in-context demonstrations? For LLMs, it is still a common practice to select a subset $\mathcal{S}$ consisting of examples that are diverse or semantically similar to the test sample as demonstrations, \ie\ considering the diversity or similarity of $\mathcal{S}$. To investigate whether one dimension is consistently better than the other one across different tasks, we conduct some pilot experiments on CommonsenseQA \citep{commonsenseqa}, BoolQ \citep{clark-etal-2019-boolq} and AGNews \citep{zhang2015character}. Specifically, we randomly sample $100$ examples from the original test set for experiments and conduct $4$-shot learning using 
GPT-3.5 (gpt-3.5-turbo).

Following \citet{zhang2022automatic}, we use Sentence-BERT \citep{reimers-gurevych-2019-sentence} to encode all samples. For each test sample, the Similar-ICL method selects the top-$4$ similar training data based on cosine similarity while the Random-ICL method randomly samples $4$ training examples as few-shot demonstrations. Inspired by \citet{cot_wei_sc}, we apply \textit{self-consistency} with $3$ decoding paths (temperature 0.7) to Similar-ICL (named \textbf{Similar-ICL-Consistency}) and run Random-ICL $3$ times before majority voting (named \textbf{Random-ICL-Voting}) to improve the robustness. 

The results of different methods on four datasets are reported in \Cref{table:pilot_exp}. We can observe that the diversity dimension outperforms the similarity dimension on CommonsenseQA while the similarity dimension is superior to the diversity dimension on BoolQ and AGNews. Therefore, the optimal dimension for selecting demonstration examples is task-specific. Thus, it is unreasonable to claim that one dimension is consistently better than the other one in ICL demonstration selection.

Intuitively, semantically similar examples can help the model correctly answer the test query as they might share similar input-output patterns with the test sample which could unleash GPT-3.5's power of text generation. To further understand why the similarity dimension underperforms the diversity dimension on CommonsenseQA, we present a case study in \Cref{tab:sim_error}. We can see that the answer of the final demonstration example extracted by Similar-ICL-Consistency, \ie\ `most buildings' is also in the options list of the test sample, which misleads the decision process of the model, leading to a wrong answer. In addition, the selected demonstrations might not include enough important information as high similarity also results in redundancy.

Considering the strengths and weaknesses of both dimensions, we aim to design a method that can select demonstration examples that are diverse (minimizing misleading information) but still strongly correlated with the test sample, which is introduced in the next section.

\section{Iterative Demonstration Selection} \label{sec:ides_method}

\begin{figure*}[t]
  \begin{center}
   \includegraphics[width=0.92\textwidth]{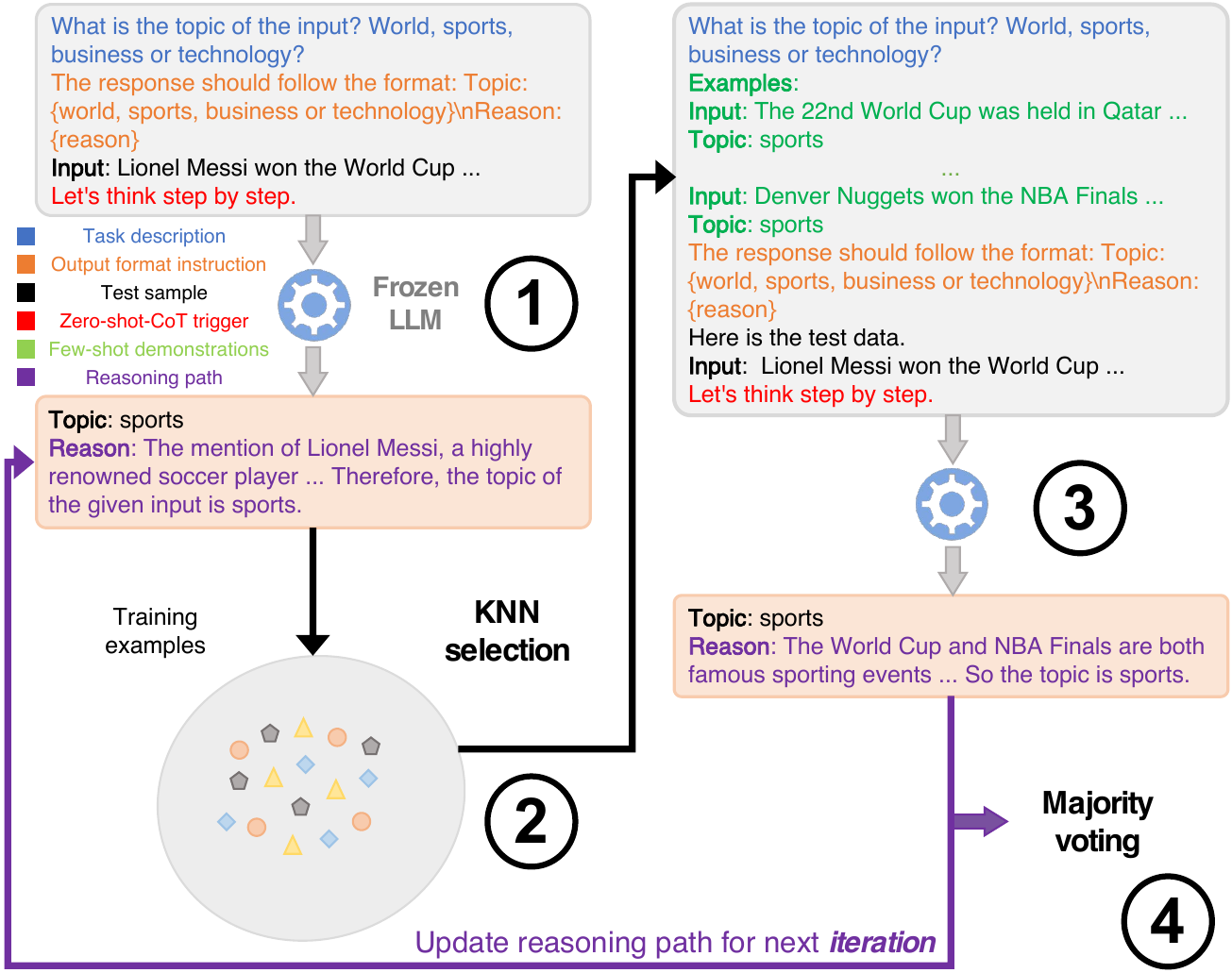}
  \end{center}
  \caption{Illustration of our proposed Iterative Demonstration Selection (\ides). \ides\ first applies Zero-shot-CoT to the test sample to obtain a reasoning path, which is then used to select few-shot demonstrations from training examples through KNN. The selected demonstration examples are prepended to the test sample for ICL. To obtain the new reasoning path for extracting another set of demonstrations in the next iteration, an instruction for output format is inserted before the test sample. After several iterations, \ides\ uses majority voting to obtain the final result.}
  \label{fig:method}
\end{figure*}

Based on the observations and considerations in \Cref{sec:two_dimensions}, we introduce Iterative Demonstration Selection (\ides) for ICL demonstration selection by leveraging \emph{how the test sample is answered} (see \Cref{fig:method} for an illustration).
Intuitively, the demonstrations that are similar to the \textit{reason} for answering a sample are strongly correlated with this sample. Therefore, we propose to incorporate zero-shot chain-of-thought reasoning (Zero-shot-CoT) into \ides\ to iteratively select demonstration examples that are diverse but still have a strong correlation with the test sample.

Specifically, for each test sample $\hat x_i$, \ides\ mainly consists of four steps:

\begin{enumerate}[leftmargin=*]
    \item  We apply \textbf{Zero-shot-CoT}, \ie\ ``Let's think step by step.'' to the test sample $\hat x_i$ before selecting demonstrations to obtain a reasoning path $\mathrm{R}$.
    \item The \textbf{reasoning path} $\mathrm{R}$ is then used to select top-$k$ ($k$ is the number of shot) most semantically similar training examples $\{(x_1,y_1),...,(x_k,y_k)\}$ as few-shot demonstrations. We use Sentence-BERT \citep{reimers-gurevych-2019-sentence} to encode the reasoning path $\mathrm{R}$ and training examples to obtain the contextual representations and use cosine similarity to measure the similarity between representations.
    \item The selected $k$ training examples $\{(x_1,y_1),...,(x_k,y_k)\}$ are then prepended to the test sample $\hat x_i$ for ICL. During inference, we ensure that the generated answer $\hat{\mathrm{A}}$ is accompanied by its corresponding reasoning path $\hat{\mathrm{R}}$ through designed prompts, \eg\, ``The response should follow the format: Topic: \{world, sports, business or technology\}\textbackslash nReason: \{reason\}''. Note that \textbf{Zero-shot-CoT} is also applied in this step to improve the quality of generated reasoning paths. After ICL, we go back to Step 2 for \emph{iterations} using the \emph{new} reasoning path $\hat{\mathrm{R}}$.
    \item After $q$ rounds of iterations between Step 2 and 3, we adopt \textbf{majority voting} on all $\hat{\mathrm{A}}$ to obtain the final result $\hat{\mathrm{A}}_{final}$. 
\end{enumerate}

\begin{algorithm}[t!]
\caption{\small Selection process of \ides}
\small 
\begin{algorithmic}[1]
    \Require Training set $\mathcal{D}_{\text{train}}$, test set $\mathcal{D}_{\text{test}}$, $\text{LLM}_\theta$, number of demonstrations $k$, number of iterations $q$ and answer set $\hat{\mathrm{A}}_{all} = \emptyset$
    \State \textsc{Encode} all samples in $\mathcal{D}_{\text{train}}$ using Sentence-BERT \Comment{\blue{Encode training set}}
    \For{$\hat x_i$ in $\mathcal{D}_{\text{test}}$}
    \State \textsc{Apply} Zero-shot-CoT to $\hat x_i$ to obtain the reasoning path $\mathrm{R}$ \Comment{\blue{Zero-shot-CoT}}
    \For{$j = 1, \ldots, q$}
        \State \textsc{Encode} $\mathrm{R}$ using Sentence-BERT    \Comment{\blue{Encode reasoning path}}
        \State \textsc{Use} $\mathrm{R}$ to select top-$k$ most similar examples $\mathcal{S} = \{(x_1,y_1),...,(x_k,y_k)\}$ from $\mathcal{D}_{\text{train}}$ as demonstrations \Comment{\blue{KNN selection}}
        \State ($\hat{\mathrm{A}}$, $\hat{\mathrm{R}}$) = $\text{LLM}_\theta (S, \hat x_i)$ \Comment{\blue{ICL with Zero-shot-CoT}}
        \State $\mathrm{R} = \hat{\mathrm{R}}$, $\hat{\mathrm{A}}_{all} = \hat{\mathrm{A}}_{all}  \cup \{\hat{\mathrm{A}}\}$   \Comment{\blue{Update reasoning path and answer set}}
    \EndFor
    \State \textsc{Adopt} majority voting for $\hat{\mathrm{A}}_{all}$ to obtain the final result $\hat{\mathrm{A}}_{final}$ for the test sample $\hat x_i$ \Comment{\blue{Majority voting}}
    \EndFor
\end{algorithmic}
\label{alg:whole_process} 
\end{algorithm}

Obviously, the selected demonstration examples are strongly correlated with the original test sample, \ie\ achieving similarity, as they are selected by the generated reasoning paths (see \Cref{sec:correlation_reasoning_path} for quantitative analysis of reasoning paths). And they can be different during iterations to achieve diversity because the reasoning paths vary in different iterations. Note that there is \emph{no} reasoning path in few-shot demonstrations (as shown in the \textcolor{applegreen}{green} part in \Cref{fig:method}). The reasoning path only exists in the output of LLMs. 

In addition, we illustrate the whole selection process in \Cref{alg:whole_process} and show the instructions and input formats of different types of tasks for ICL in \Cref{ins_inputf_datasets}.

\begin{table*}[t]    
    \centering
    \resizebox{0.95\linewidth}{!}{
    \begin{tabular}{
        l @{\hspace{2em}}
        ccccccc
        }
        \toprule
            \textbf{Method} & BoolQ & GSM8K & MATH & CommonsenseQA & LogiQA &  AGNews & Average \\
        \midrule
        Vote-$k$    & $86.7_{\pm 0.7}$ & $76.5_{\pm 0.5}$ & $35.7_{\pm 0.2}$ & $75.2_{\pm 0.3}$    & $45.4_{\pm 0.3}$ & $88.1_{\pm 1.2}$ & $67.9_{\pm 0.2}$ \\
        MMR           & $86.4_{\pm 0.8}$ & $75.5_{\pm 0.7}$ & $34.8_{\pm 0.3}$ & $74.9_{\pm 0.2}$    & $44.7_{\pm 0.3}$ & $87.6_{\pm 1.1}$  & $67.3_{\pm 0.3}$ \\
        G-fair-Prompting    & $84.8_{\pm 0.7}$ & $76.9_{\pm 0.6}$ & $34.6_{\pm 0.3}$ & $75.5_{\pm 0.3}$    & $43.8_{\pm 0.4}$ & $88.9_{\pm 1.0}$ & $67.4_{\pm 0.2}$ \\
        Skill-KNN           & $85.9_{\pm 0.5}$ & $76.5_{\pm 0.3}$ & $35.1_{\pm 0.2}$ & $75.2_{\pm 0.2}$    & $44.6_{\pm 0.2}$ & $88.7_{\pm 0.9}$  & $67.7_{\pm 0.1}$ \\
        Top-$k$-Consistency & $87.1_{\pm 0.2}$ & $76.1_{\pm 0.5}$ & $35.6_{\pm 0.3}$ & $74.5_{\pm 0.2}$    & $45.7_{\pm 0.4}$ & $89.3_{\pm 0.8}$ & $68.1_{\pm 0.1}$ \\
        Random-Voting       & $87.3_{\pm 0.6}$ & $75.6_{\pm 0.4}$ & $35.4_{\pm 0.1}$ & $77.0_{\pm 0.2}$    & $45.1_{\pm 0.3}$ & $87.0_{\pm 1.6}$ & $67.9_{\pm 0.2}$ \\
        Cluster-Voting      & $86.4_{\pm 0.7}$ & $76.8_{\pm 0.3}$ & $34.9_{\pm 0.4}$ & $76.5_{\pm 0.3}$    & $44.1_{\pm 0.3}$ & $86.8_{\pm 1.2}$ & $67.6_{\pm 0.3}$  \\
        \ides               & $\mathbf{87.8}_{\pm 0.8}$             & $\mathbf{78.5}_{\pm 0.4}$ & $\mathbf{37.5}_{\pm 0.2}$ & $\mathbf{78.1}_{\pm 0.1}$      & $\mathbf{46.9}_{\pm 0.2}$ & $\mathbf{89.8}_{\pm 0.8}$      & $\mathbf{69.8}_{\pm 0.1}$ \\
        \bottomrule
    \end{tabular}
    }
    \caption{ Accuracy ($\%$) of different methods on 6 datasets.  \textbf{Bold} indicates the best result. \ides\ is consistently better than all previous baselines. 
    }
    \label{tab:main-result}
\end{table*}

\section{Experiments}

In this section, we first describe the tasks and datasets, and then introduce methods compared in our work. Finally, we present the experimental results.

\subsection{Experimental Setup}

\paragraph{Tasks and Datasets} We mainly investigate 6 different datasets covering 5 representative task categories: mathematical reasoning (GSM8K \citep{cobbe2021training} and MATH \citep{hendrycks2021measuring}), commonsense reasoning (CommonsenseQA \citep{commonsenseqa}), logical reasoning (LogiQA \citep{liu2020logiqa}), question answering (BoolQ \citep{clark-etal-2019-boolq}) and topic classification (AGNews \citep{zhang2015character}). For each dataset, we randomly sample at most 10000 examples from the original training set as $\mathcal{D}_{\text{train}}$ and at most 2000 test examples as $\mathcal{D}_{\text{test}}$ for evaluating the performance of selected demonstrations. The detailed information of different datasets is shown in \Cref{dataset_info}. To reduce the randomness, we run every experiment five times with different random seeds (resulting in different training and test samples if not using the whole set) and report the average results. Without specification, we use $k$ = 4 number of demonstrations following \citet{wang2023large} and set the number of iterations $q$ to $3$.

\paragraph{Methods Compared} We mainly use 
GPT-3.5 (gpt-3.5-turbo) as the LLM and compare our \ides\ with the following methods in the experiments for selecting ICL demonstrations:

\begin{itemize}[leftmargin=*]
    \item \textbf{Top-$k$-Consistency} \citep{liu-etal-2022-makes} selects the \textit{top-$k$} semantically similar examples from the training set $\mathcal{D}_{\text{train}}$ as demonstrations for each test sample and applies \textit{self-consistency} \citep{cot_wei_sc} with $q$ decoding paths (temperature 0.7) to match the number of iterations. Following \citet{zhang2022automatic}, all samples are encoded by Sentence-BERT \citep{reimers-gurevych-2019-sentence} to obtain contextual representations for calculating the cosine similarity.
    \item \textbf{Random-Voting} randomly selects $k$ examples from $\mathcal{D}_{\text{train}}$ as few-shot demonstrations for every test sample and runs experiments $q$ times before majority voting.
    \item \textbf{Cluster-Voting} partitions $\mathcal{D}_{\text{train}}$ into $k$ clusters and selects a representative example from each cluster to form demonstrations. Following \citet{zhang2022automatic}, we choose the sample closest to the centroid in each cluster as the representative example. Same as Random-Voting, after running experiments $q$ times, Cluster-Voting adopts majority voting to obtain the final result.
\end{itemize}

Besides, we also compare \ides\ with several latest ICL demonstration selection approaches: \textbf{Vote-$k$} \citep{su2023selective}, \textbf{MMR} \citep{ye-etal-2023-complementary}, \textbf{G-fair-Prompting} \citep{ma2023fairness} and \textbf{Skill-KNN} \citep{an2023skill}
(see \Cref{sec:detail_baseline} for more details of baselines). Similar to Top-$k$-Consistency, we apply self-consistency to these baselines to match the number of iterations $q$. Note that we find that simultaneously generating answers and reasoning paths can improve the ICL performance in general even if the target task is not a reasoning task in the conventional sense, \eg\ topic classification. Therefore, we apply the same prompt, \eg\ ``The response should follow the format: Topic: \{world, sports, business or technology\}\textbackslash nReason: \{reason\}'', and \textit{Zero-shot-CoT} to baseline methods.

\begin{table}[t]
\centering
\setlength\tabcolsep{3pt}
\scalebox{0.68}{
\begin{tabular}{l|c|c|c}
\toprule
\multirow{1}{*}{} & \multicolumn{1}{c}{Top-$k$-Consistency}    & \multicolumn{1}{|c}{\ides}   & \multicolumn{1}{|c}{Random-Voting}   \\
\midrule

  Average Similarity Score &   0.68 & 0.48  & 0.32
\\

\bottomrule
\end{tabular}
}
\caption{Average similarity scores between test examples and the corresponding selected demonstrations of three methods (Top-$k$-Consistency, \ides\ and Random-Voting).} 
\label{table:ave_sim_score}
\end{table}

\begin{figure}[t]
    \centering
    \includegraphics[width=0.45\textwidth]{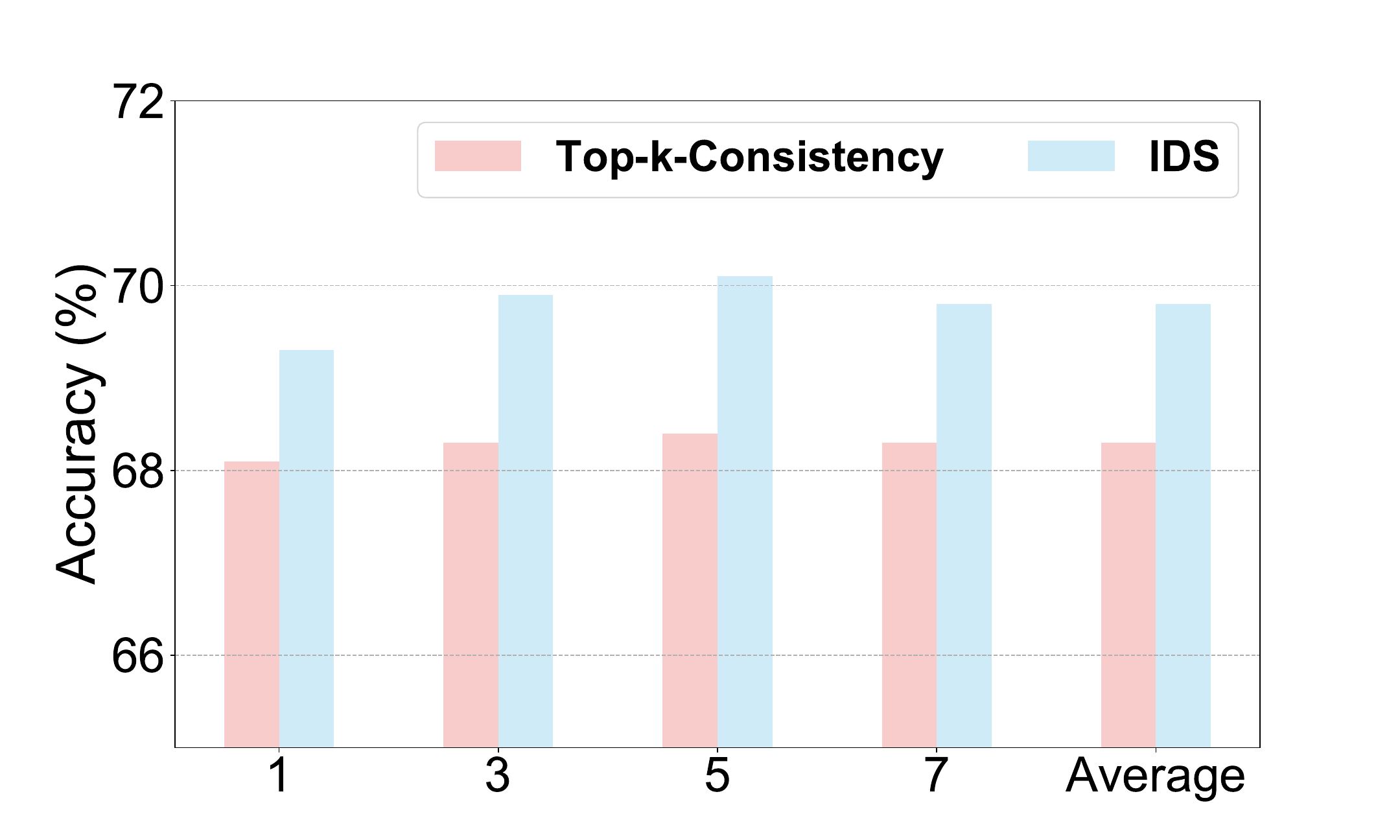}
    \caption{Accuracy ($\%$) of Top-$k$-Consistency and \ides\ with different numbers of reasoning paths or iterations.}
    \label{fig:diff_round}
\end{figure}

\subsection{Main Results}

\Cref{tab:main-result} shows the average performance scores of different methods on all investigated datasets. From the results, we can observe that

\noindent $\bullet~$ Our proposed \ides\ consistently outperforms previous baselines on all datasets with a negligible increase in API request cost (Zero-shot-CoT in the first step), which demonstrates that our method can indeed effectively and efficiently select better ICL demonstration examples by incorporating the reason for answering the test query. 

\noindent $\bullet~$ On average, \ides\ yields about 1.7$\%$ performance boost compared with the best baseline Top-$k$-Consistency as it can fully leverage the merits of both selection dimensions (diversity and similarity). While the performance gain on a few simple benchmarks looks somewhat small (because the baseline results are already pretty high, \eg\ the baseline performance of BoolQ and AGNews is above 85$\%$), \ides\ performs much better than baselines on more complex tasks. For example, \ides\ can bring an average relative improvement of about 4$\%$ on mathematical reasoning tasks compared with Top-$k$-Consistency.

To delve deeper into how different dimensions are leveraged in selected demonstrations, we report the average similarity scores between test samples and the corresponding demonstrations of different methods in \Cref{table:ave_sim_score}. Specifically, we randomly select 500 test examples for each dataset and use Sentence-BERT to obtain contextual representations for calculating similarity scores. We can see that the average similarity score of \ides\ is between that of Top-$k$-Consistency and Random-Voting, indicating that it can indeed strike a balance between two selection dimensions (see \Cref{sec:demon_diversity} for more analysis on the diversity of the selected demonstration examples).

\subsection{Analysis}

\paragraph{Different Numbers of Iterations} Our experiments and analysis so far use $q=3$ iterations. To verify whether the performance gain of \ides\ is consistent across different numbers of iterations, we conduct controlled experiments with $q = \{1, 5, 7\}$. The average results of the 6 datasets with a randomly selected seed are reported in \Cref{fig:diff_round}. \ides\ consistently outperforms the best baseline Top-$k$-Consistency with different $q$ (even $q = 1$, \ie\ without voting), emphasizing the importance of rationales in selecting demonstration examples. Interestingly, the performance of ICL does not always improve with the number of iterations, which might be because increased iterations can also lead to unnecessary noise; we provide an in-depth analysis in \Cref{sec:noise_increased_iterations}.

\begin{table}[t]

\centering
\setlength\tabcolsep{3pt}
\scalebox{0.90}{
\begin{tabular}{l|c|c}
\toprule
\multirow{1}{*}{} & \multicolumn{1}{c}{GPT-3.5}    & \multicolumn{1}{|c}{GPT-4}     \\
\midrule

Top-$k$-Consistency   &   68.3 & 73.9 
\\
\ides &   \textbf{69.9} & \textbf{75.4} 
 \\

\bottomrule
\end{tabular}
}
\caption{Accuracy ($\%$) of Top-$k$-Consistency and \ides\ with different LLMs (GPT-3.5 and GPT-4). For GPT-4, we randomly sample $200$ test examples per dataset due to the high cost.} 
\label{table:diff_model}
\end{table}

\paragraph{Robustness to Model Types} To demonstrate the robustness of \ides\ to model types, we conduct controlled experiments with GPT-4. 
Specifically, we randomly select one seed and sample $200$ test examples per dataset for experiments due to the expensive cost. From the average results shown in \Cref{table:diff_model}, we can observe that \ides\ still achieves better performance than Top-$k$-Consistency when using GPT-4 as the LLM, showing its robustness to different LLMs.

\paragraph{Generalization to Open-source LLMs} To better verify the generalization ability of \ides, we use vLLM \citep{kwon2023efficient} to serve Llama-2-chat models \citep{touvron2023llama} for experiments and compare \ides\ with Top-$k$-Consistency on two datasets: BoolQ and GSM8K. We randomly sample 500 test examples for experiments and report the results in \Cref{table:other_openllms}, which demonstrates that \ides\ can successfully generalize to open-source LLMs of different sizes.

\begin{table}[t]

\centering
\setlength\tabcolsep{3pt}
\scalebox{0.82}{
\begin{tabular}{lcccccc}
\toprule
\multirow{2}{*}{} & \multicolumn{3}{c}{BoolQ} & \multicolumn{3}{c}{GSM8K}  \\
\cmidrule(lr){2-4}  \cmidrule(lr){4-7} & \multicolumn{1}{c}{7B}    &  \multicolumn{1}{c}{13B} &
\multicolumn{1}{c}{70B}    &  \multicolumn{1}{c}{7B} & \multicolumn{1}{c}{13B}    &  \multicolumn{1}{c}{70B}   \\
\midrule

 Top-$k$-Consistency  & 77.1 & 81.3 &   84.2 & 14.6 & 24.8    & 49.6
\\
  \ides & \textbf{78.5} &  \textbf{82.2} &   \textbf{85.4} &   \textbf{16.6} & \textbf{27.1} & \textbf{51.4}
\\

\bottomrule
\end{tabular}
}
\caption{Accuracy ($\%$) of different methods with Llama-2-chat models.} 
\label{table:other_openllms}
\end{table}

\begin{figure}[t]
  \begin{center}
   \includegraphics[width=0.95\columnwidth]{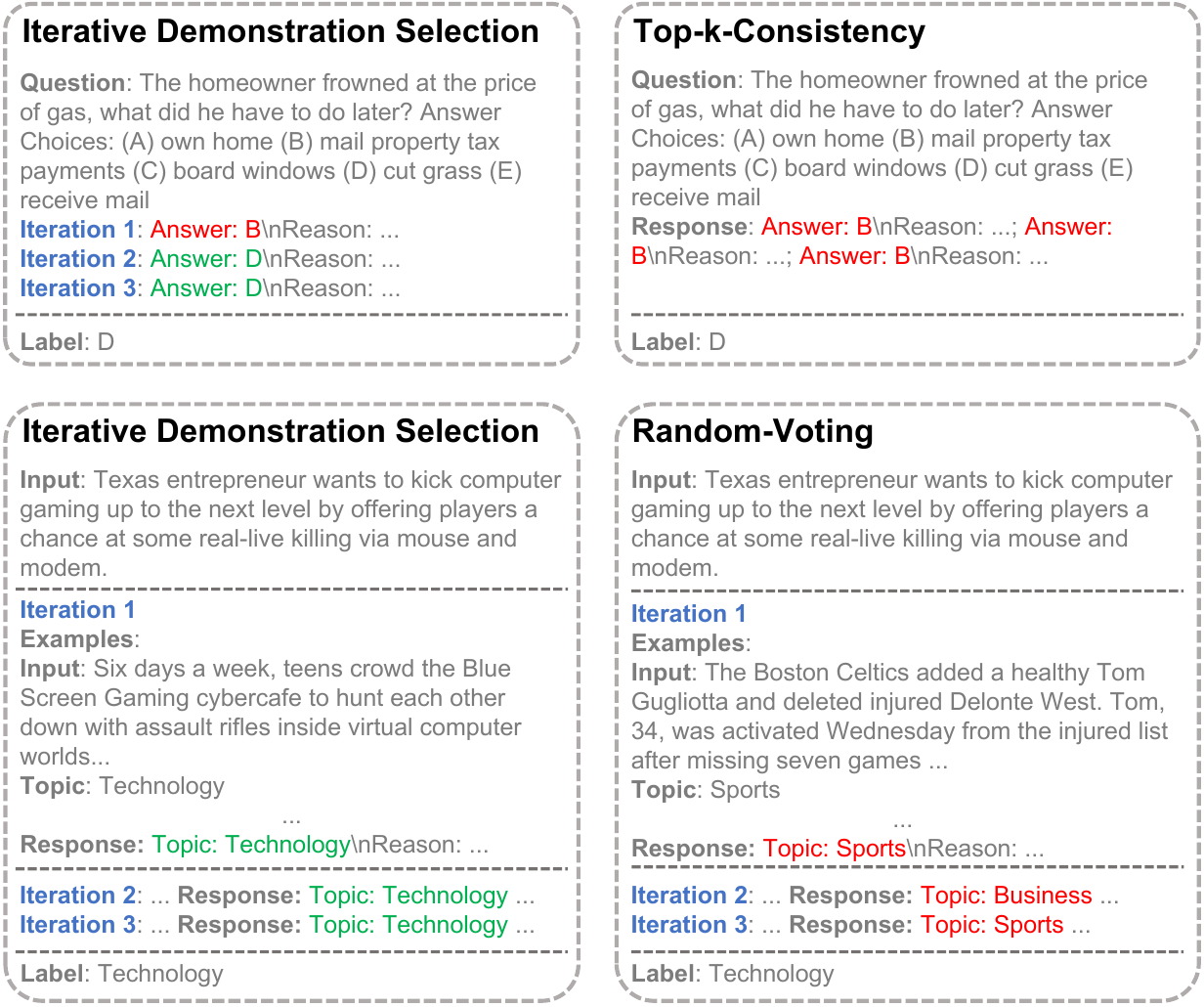}
  \end{center}
  \caption{Several case studies of model responses. We color correct outputs in \textcolor{applegreen}{green}, and wrong outputs in \textcolor{red}{red}.}
  \label{fig:case_study}
\end{figure}

\paragraph{Case Study} To further understand the advantage of \ides, we show several cases in \Cref{fig:case_study}. As shown in the upper part of the figure, \ides\ can iteratively select more diverse demonstration examples than Top-$k$-Consistency which may be able to correct errors from previous iterations. Compared with Random-Voting, \ides\ can find examples that share more similar input-output patterns with the test sample to induce the LLM to generate correct answers (the lower part of the figure).

In addition, we show the results with different numbers of demonstrations, the robustness of \ides\ to different embedding models and Zero-shot-CoT triggers in Appendix \ref{sec:diff_num_demons} $\sim$ \ref{sec:robustness_trigger}, respectively.

\section{Conclusion}

In this work, we have introduced Iterative Demonstration Selection (\ides) that can iteratively select examples that are diverse but still strongly correlate with the test sample as demonstrations to improve the performance of in-context learning (ICL) by leveraging the rationale for answering the test sample. Extensive experimental results and analysis show that \ides\ can consistently outperform previous ICL demonstration selection baselines.

\section*{Limitations}

This work has several limitations. First, due to the inference cost of ChatGPT,
we do not conduct experiments on the entire test set. Besides, we include 6 datasets covering 5 different task types in this work. A further improvement could be to explore more diverse types of tasks.

\bibliography{anthology,custom}

\appendix

\section{Appendix}

\subsection{Instructions and Input Formats of Different Tasks} \label{ins_inputf_datasets}

\begin{figure*}[t]
  \begin{center}
   \includegraphics[width=1.0\textwidth]{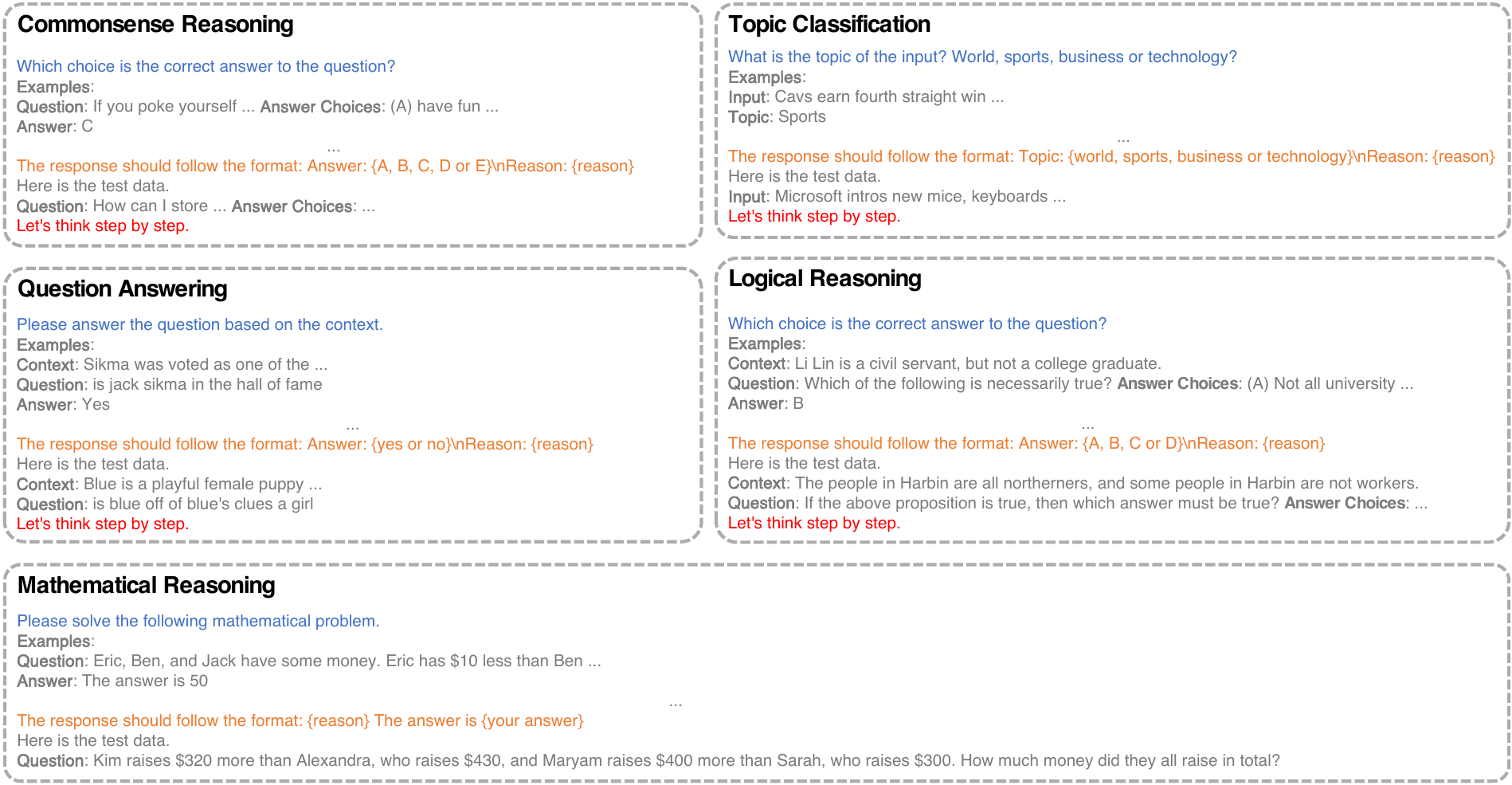}
  \end{center}
  \caption{Instructions and input formats of five different categories of tasks (topic classification, question answering, commonsense reasoning, logical reasoning, and mathematical reasoning) for ICL. For Zero-shot-CoT in the first step of \ides, there is no demonstration example and the instruction ``Here is the test data.''.}
  \label{fig:prompt}
\end{figure*}

We show the instructions and input formats of different types of tasks for in-context learning in \Cref{fig:prompt}.

\subsection{Datasets Information} \label{dataset_info}

\begin{table*}[t]
    
    \centering
    \resizebox{0.95\linewidth}{!}{
    \begin{tabular}{
        l @{\hspace{2em}}
        cccccc
        }
        \toprule
            \textbf{} &  BoolQ & GSM8K &  MATH & CommonsenseQA  & LogiQA &  AGNews   \\
        \midrule
        \# \textbf{Training Samples} & 9427 (full) & 7473(full) &  5000 & 9741 (full)  & 7376(full) &  10000  \\
        \# \textbf{Test Samples} & 2000 & 1000 &  1000 & 1221 (full) & 500 &  1000  \\
        \bottomrule
    \end{tabular}
    }
    \caption{Deailed information of different datasets. \# refers to `the number of' and `full' means the whole set. Note that different random seeds do not result in different samples if the whole set is used.
    }
\label{tab:information}
\end{table*}

We show the detailed information of different datasets in \Cref{tab:information}.

\subsection{Details of Baselines} \label{sec:detail_baseline}

In this work, we compare \ides\ with the following latest ICL demonstration selection approaches:

\begin{itemize}[leftmargin=*,topsep=3pt,itemsep=3pt,parsep=0pt]

\item \textbf{Vote-$k$} \citep{su2023selective} is an unsupervised, graph-based selective annotation method used for selecting and annotating diverse, representative examples. The annotated examples then serve as a pool for demonstration retrieval.
\item \textbf{MMR} \citep{ye-etal-2023-complementary} proposes a maximal marginal relevance-based approach for demonstration selection.
\item \textbf{G-fair-Prompting} \citep{ma2023fairness} leverages greedy search to select the example with the highest fairness score at each step.
\item \textbf{Skill-KNN} \citep{an2023skill} generates skill-based descriptions for test queries and then uses these descriptions to select similar examples as demonstrations.

\end{itemize}

\subsection{Measure of Reasoning Path Correlation} \label{sec:correlation_reasoning_path}

\begin{table}[t]
\centering
\setlength\tabcolsep{3pt}
\scalebox{0.75}{
\begin{tabular}{l|c|c|c}
\toprule
\multirow{1}{*}{} & \multicolumn{1}{c}{$\text{score}_\text{{reason}}$}    & \multicolumn{1}{|c}{$\text{score}_\text{{random}}$}   & \multicolumn{1}{|c}{$\text{score}_\text{{similar}}$}   \\
\midrule

Average Similarity Score &   0.59 & 0.32  & 0.68
\\

\bottomrule
\end{tabular}
}
\caption{Comparison between different average similarity scores.} 
\label{table:reasoning_path_correlation}
\end{table}

We report the average similarity score between test samples and the corresponding generated reasoning paths ($\text{score}_\text{{reason}}$), the average similarity score between test samples and randomly selected training examples ($\text{score}_\text{{random}}$), and the average similarity score between test samples and the most similar training examples ($\text{score}_\text{{similar}}$) in \Cref{table:reasoning_path_correlation}. For each dataset, we randomly select 500 test samples and use Sentence-BERT for similarity calculation. We can observe that $\text{score}_\text{{reason}}$ is slightly worse than $\text{score}_\text{{similar}}$ and much higher than $\text{score}_\text{{random}}$, indicating that the generated reasoning path is indeed strongly correlated with the test sample.

\subsection{Analysis on Demonstration Diversity} \label{sec:demon_diversity}

\begin{table}[t]
\centering
\setlength\tabcolsep{3pt}
\scalebox{0.68}{
\begin{tabular}{l|c|c}
\toprule
\multirow{1}{*}{} & \multicolumn{1}{c}{Top-$k$-Consistency}    & \multicolumn{1}{|c}{\ides} \\
\midrule

Average Pairwise Similarity &   0.55 & 0.39 
\\

\bottomrule
\end{tabular}
}
\caption{Comparison of average pairwise similarity scores of demonstrations selected by different methods.} 
\label{table:new_metrics_scores}
\end{table}

In addition to the average similarity score between test samples and demonstrations, we further calculate the following metrics for \ides\ and Top-$k$-Consistency:
\begin{align}
 Q_{S} = \sum_{1 \le i < j \le |S|} g(S_i, S_j) / C(|S|, 2) \label{eq:new_metrics}
\end{align}
where $S$ is the set of the selected demonstration examples, and $g$ is the function of measuring similarity. $Q$ calculates the average pairwise similarity score of the demonstrations, which can be used to reflect whether they are diverse from each other. As can be seen from the results in \Cref{table:new_metrics_scores}, the average pairwise similarity score of \ides\ is much lower than that of Top-$k$-Consistency, verifying the diversity of demonstration examples selected by \ides.

\subsection{Noise Caused by Increased Iterations} \label{sec:noise_increased_iterations}

\begin{table}[t]
\centering
\setlength\tabcolsep{3pt}
\scalebox{0.80}{
\begin{tabular}{l|c|c}
\toprule
\multirow{1}{*}{\textbf{Iteration}} & \multicolumn{1}{c}{5}    & \multicolumn{1}{|c}{7} \\
\midrule

$\text{Prop}_{\text{pre}}$ &   31.9$\%$ & \textbf{60.4}$\%$ 
\\
\midrule
$\text{Prop}_{\text{pre}}^{\text{wrong}}$ &   13.1$\%$ & \textbf{38.7}$\%$ 
\\
\bottomrule
\end{tabular}
}
\caption{Comparison between different iterations.} 
\label{table:noise_reason_analysis}
\end{table}

As observed from \Cref{fig:diff_round}, the performance of ICL does not always improve with the number of iterations. We speculate that this is because too many iterations may also lead to unnecessary noise. As the number of iterations increases, the demonstrations selected in the latest iteration are more likely to have been chosen in previous iterations. Therefore, if these demonstrations result in wrong answers in previous iterations, these errors may be propagated to later iterations, \ie\ unnecessary noise caused by increased iterations. To better verify our hypothesis, we calculate \Ni the proportion of demonstrations selected in iteration 5 or 7 that were also chosen in previous iterations ($\text{Prop}_{\text{pre}}$), and \Nii the proportion of demonstrations selected in iteration 5 or 7 that were chosen in previous iterations and resulted in wrong answers ($\text{Prop}_{\text{pre}}^{\text{wrong}}$). We can see from \Cref{table:noise_reason_analysis} that the results of the 7th iteration are much higher than those of the 5th iteration, indicating the correctness of our claim.

\subsection{Different Numbers of Demonstrations} \label{sec:diff_num_demons}

\begin{table}[t]
\centering
\setlength\tabcolsep{3pt}
\scalebox{0.90}{
\begin{tabular}{l|c|c|c|c}
\toprule
\multirow{1}{*}{} & \multicolumn{1}{c}{2}    & \multicolumn{1}{|c}{4}       & \multicolumn{1}{|c}{6}                    & \multicolumn{1}{|c}{8}              \\
\midrule

Top-$k$-Consistency   &  68.0 & 68.3 & 68.5 & 68.4\\
\ides &  \textbf{69.4} & \textbf{69.9} & \textbf{69.9} & \textbf{69.7} \\

\bottomrule
\end{tabular}
}
\caption{Accuracy ($\%$) of Top-$k$-Consistency and \ides\ with different numbers of demonstrations $k$.} 
\label{table:diff_shot}
\end{table}

While we use $k=4$ demonstration examples for all experiments, we also evaluate the effectiveness of \ides\ with different $k$. We randomly choose one seed for experiments and report the average results of the 6 datasets in \Cref{table:diff_shot}. We can see that \ides\ consistently outperforms Top-$k$-Consistency with different numbers of demonstrations. In addition, more demonstrations do \textbf{not} guarantee better ICL performance, which is consistent with the observation in \citet{wang2023large}.

\begin{table}[t]
\small
\centering
\setlength\tabcolsep{3pt}
\scalebox{0.95}{
\begin{tabular}{l|c|c|c}
\toprule
\multirow{1}{*}{} & \multicolumn{1}{c}{BoolQ}    & \multicolumn{1}{|c}{CommonsenseQA}   & \multicolumn{1}{|c}{GSM8K}   \\
\midrule

 Top-$k$-Consistency  &   86.0 & 75.4  & 75.8
\\
  \ides &   \textbf{87.2} & \textbf{78.0}  & \textbf{77.6}
\\

\bottomrule
\end{tabular}
}
\caption{Accuracy ($\%$) of different methods with OpenAI embedding model (text-embedding-ada-002) on three datasets.} 
\label{table:other_embedding}
\end{table}

\subsection{Robustness to Embedding Models} \label{sec:robustness_encoder}

Instead of using Sentence-BERT, we also explore adopting the OpenAI embedding model (text-embedding-ada-002) as the encoder. Specifically, we conduct experiments on 3 datasets: BoolQ, CommonsenseQA and GSM8K. For each dataset, we randomly sample 500 test examples and compare \ides\ with the baseline Top-$k$-Consistency. The results reported in \Cref{table:other_embedding} demonstrate IDS's robustness to different embedding models.

\begin{table}[t]
\small
\centering
\setlength\tabcolsep{3pt}
\scalebox{0.95}{
\begin{tabular}{l|c|c|c}
\toprule
\multirow{1}{*}{} & \multicolumn{1}{c}{Default}    & \multicolumn{1}{|c}{Trigger1}   & \multicolumn{1}{|c}{Trigger2}   \\
\midrule

\ides  &   70.1 & 70.3  & 70.0
\\

\bottomrule
\end{tabular}
}
\caption{Accuracy ($\%$) of \ides\ with different Zero-shot-CoT triggers.} 
\label{table:other_trigger}
\end{table}

\subsection{Robustness to Zero-shot-CoT Triggers} \label{sec:robustness_trigger}

To verify the robustness of \ides\ to Zero-shot-CoT triggers, we conduct controlled experiments with two new triggers: ``Let's work this out in a step by step way to be sure we have the right answer.'' (Trigger1) and ``Let's solve this problem step by step'' (Trigger2). Specifically, we randomly sample 500 test examples per dataset for experiments and report the average results in \Cref{table:other_trigger}, which demonstrates that \ides\ is indeed robust to different Zero-shot-CoT triggers.

\end{document}